\documentclass[letterpaper, 10 pt, conference]{ieeeconf}
\IEEEoverridecommandlockouts
\usepackage{epsfig} %% for loading postscript figures
\usepackage{amsmath}
\usepackage{multirow}
\usepackage{tabularx}
\usepackage{color}
\usepackage{algorithmicx}
\usepackage{algpseudocode}
\usepackage{bm}
\usepackage{amsfonts}
\usepackage{graphics}
\usepackage{url}
\usepackage{siunitx}
\usepackage{mathrsfs}
\usepackage{soul}

\soulregister{\cite}7
\soulregister{\citep}7
\soulregister{\citet}7
\soulregister{\ref}7
\soulregister{\pageref}7
\soulregister{\uppercase\expandafter}7
\sethlcolor{yellow}

\title{\LARGE \bf
Risk-aware Trajectory Sampling for Quadrotor Obstacle Avoidance in Dynamic Environments
}

\author{Gang~Chen, Peng Peng, Peihan Zhang, and Wei~Dong% <-this % stops a space
\thanks{Gang~Chen, Peng Peng, Peihan Zhang, and Wei~Dong are with
        the State Key Laboratory of Mechanical System and Vibration,
        School of Mechanical Engineering, Shanghai Jiaotong University,
        Shanghai, 200240, China.
        }%
\thanks{ Corresponding author: Wei~Dong.
        {\tt\small dr.dongwei@sjtu.edu.cn}
        }
        }

\begin{document}
\maketitle
\thispagestyle{empty}
\pagestyle{empty}

\begin{abstract}
Obstacle avoidance of quadrotors in dynamic environments is still a very open problem. Current works commonly leverage traditional static maps to represent static obstacles and the detection and tracking of moving objects (DATMO) method to model dynamic obstacles separately. The detection module requires pre-training, and the dynamic obstacles can only be modeled with certain shapes, such as cylinders or ellipsoids. This work utilizes the dual-structure particle-based (DSP) dynamic occupancy map to represent the arbitrary-shaped static obstacles and dynamic obstacles simultaneously, and proposes an efficient risk-aware sampling-based local trajectory planner to realize safe flights in this map. The trajectory is planned by sampling motion primitives generated in the state space. Each motion primitive is divided into two phases: a short-term phase with a strict risk limitation and a relatively long-term phase designed to avoid high-risk regions. The risk is evaluated with the predicted particle-form future occupancy status, considering the time dimension. With an approach to split from and merge to an arbitrary global trajectory, the planner can also be used in the tasks with preplanned global trajectories. Comparison experiments show that the obstacle avoidance system composed of the DSP map and our planner performs the best in dynamic environments. In real-world tests, our quadrotor reaches a speed of 6 m/s with the motion capture system and 2.5 m/s with everything running on a low-price single-board computer.

\end{abstract}


% Note that keywords are not normally used for peerreview papers.
%\begin{IEEEkeywords}
%visual navigation, end-to-end learning, sim-to-real
%\end{IEEEkeywords}

\section{Introduction}
Compared with obstacle avoidance in static environments, obstacle avoidance in dynamic environments requires the prediction of future statuses of dynamic obstacles and a planner that considers the uncertainty in the future statuses.
Current obstacle avoidance systems in dynamic environments \cite{ActiveSensing} \cite{GaofeiDynamic} \cite{ZhuhaiNew} \cite{ZhuhaiOld} of quadrotors commonly leverage DATMO to track the dynamic obstacles and describe each of them with a Gaussian distribution. The shape of the dynamic obstacles can only be modeled with certain shapes, such as cylinders or ellipsoids, and realize collision checking or collision avoidance constraints in the trajectory planner.
The static obstacles, if considered, are modeled separately with occupancy maps, such as the Octomap \cite{OctoMap}, constructed from point clouds \cite{ActiveSensing} \cite{GaofeiDynamic}.
However,  DATMO can only model the dynamic obstacles that were trained in the detector, and using cylinders or ellipsoids is too conservative for dynamic obstacles with arbitrary shapes. In addition, in dynamic environments, static occupancy maps have trail noise \cite{ActiveSensing} that can affect flight safety.

\begin{figure}
\centerline{\psfig{figure=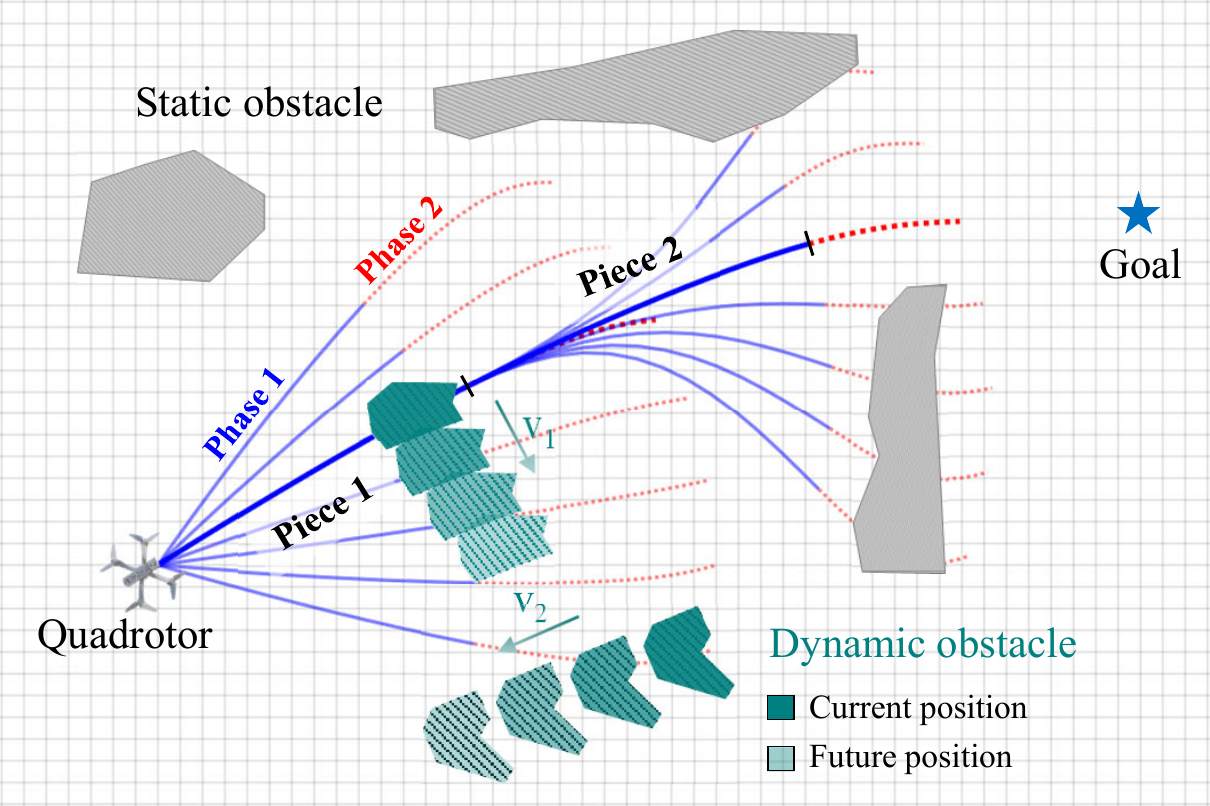, width=3.3in}}
\caption{Illustration of our planner working in a dynamic environment. The states of dynamic obstacles and static obstacles are represented and predicted simultaneously in a dynamic occupancy map \cite{OursDynamicMap}. The flight trajectory is multiple pieces. Each piece is sampled with motion primitives in the state space. Each motion primitive is divided into two phases to guarantee safety in the short term and lead to a low-risk region in the relatively long term.}
\label{Figure: Cover}
\end{figure}

In \cite{OursDynamicMap}, we proposed an efficient DSP map that models the occupancy status of static obstacles and dynamic obstacles simultaneously without using any detector. Occupancy statuses of static obstacles and dynamic obstacles with arbitrary shapes can be represented in the same form and predicted for any future moment.
This work adopts this dynamic occupancy map to model the environment. To tackle the uncertainty in predicting the future occupancy status of the obstacles and realize efficient obstacle avoidance in unknown dynamic environments, we propose a Risk-Aware Sampling-based (RAS) local trajectory planner. The risk is formed with the cardinality expectation in a risk-checking corridor and is calculated with the particle weights on the map. The flight trajectory is planned efficiently by sampling motion primitives in the state space. Each motion primitive splits into two phases that are separated by a phase time. The first phase should have a risk lower than a small threshold to be considered collision-free in the short term (the phase time). In the second phase, the risk is used as a cost to find the motion primitive that leads to a low-collision-risk region in a relatively long term, which can further enhance safety and avoid freezing. The next motion primitive then starts from the intersection of the two phases. To tackle the tasks with a preplanned global trajectory, such as the patrol and transportation tasks, we also present the algorithm to fuse the local trajectory to an arbitrary global trajectory. Experimental results show that our planner can efficiently find a safe flight trajectory in the dynamic environment. The flight speed in real-world tests is up to 6 m/s when the motion capture system provides localization and ground truth positions of the obstacles and 2.5 m/s when positioning, mapping and planning are all conducted onboard with a mini quadrotor that weighs only 320 grams.

To the best of the authors' knowledge, this is the first quadrotor obstacle avoidance system that can avoid arbitrary-shaped dynamic and static obstacles simultaneously. The contributions of this paper include:
\begin{enumerate}
  \item An efficient risk-aware sampling approach composed of two-phase motion primitives in the state space.
  \item The approach to merge the local trajectory given by our planner to an arbitrary global trajectory.
  \item A complete obstacle avoidance system that can be applied to light-weight quadrotors in dynamic environments.
\end{enumerate}

%Smaller risk in the second phase suggests a lower probability of collision .
%Two kinds of risk-checking corridors, AAA corridor and BBB corridor, are considered to perform risk .
%The occupancy status at any future time can be predicted by propagating particles.

\section{Related Work}
Obstacle avoidance is fundamental for quadrotors' autonomous navigation. Various works have addressed obstacle avoidance in static environments. The most popular pipeline is to detect obstacles with lidars or depth cameras and represent them with voxels map \cite{OctoMap} \cite{Ringbuffer}. Then a safe flight trajectory can be planned by sampling-based methods \cite{OursPlanning} \cite{ControlSpaceSample} or optimization-based methods \cite{SikangLiuCorridor}  \cite{VoxelCorridor} \cite{EgoPlanner} \cite{FastPlanner}. Sampling-based methods can search for a feasible trajectory in the map without constructing convex safety corridors \cite{SikangLiuCorridor}  \cite{VoxelCorridor} or distance fields \cite{FastPlanner}, or solving optimization problems, and thus are more computationally efficient \cite{OursPlanning}.
Sampling-based methods can further be divided into control space sampling \cite{ControlSpaceSampling} and state-space sampling \cite{OursPlanning} \cite{AggressiveSample}. Compared to control space sampling, state-space sampling samples the final states in each step, and the control input is consistent, which is favorable when the flight speed is high.

In dynamic environments, obstacle avoidance is more challenging because the future statuses of dynamic obstacles need to be predicted, and the time dimension should be considered in the planner. In addition, a faster planning speed is required to accommodate the fast-changing characteristics of dynamic environments. Current obstacle avoidance systems for quadrotors detect and track the dynamic obstacles with DATMO \cite{DATMO} methods and use sampling-based or optimization-based methods that consider collisions with the time dimension. \cite{ActiveSensing} detects dynamic obstacles with YOLO \cite{YoloV3} and tracks them with SORT \cite{SORT} and active vision. The shape of obstacles is modeled as cylinders. Trajectory planning is fulfilled by sampling in state space and collision checking with the cylinder models. In \cite{GaofeiDynamic}, the researchers use a similar detection and tracking method but model the dynamic obstacles as ellipsoids. A safe trajectory is optimized by considering the predicted position of the dynamic obstacles in their previous optimization-based planner \cite{EgoPlanner}. In both works \cite{ActiveSensing} \cite{GaofeiDynamic}, static obstacles are represented separately by voxel maps, and collision to the static obstacles is also separately considered in the collision checking algorithm or the cost function for optimization. Prediction of the future status of dynamic obstacles is usually uncertain. \cite{ActiveSensing} considers the prediction uncertainty using Gaussian distributions. \cite{ZhuhaiNew} addresses the prediction uncertainty along with the self-localization uncertainty with a planner based on model predictive control. The dynamic obstacles are detected using depth images and modeled as ellipsoids, while static obstacles are not considered. In \cite{Panther}, the dynamic obstacles are detected and tracked from point clouds. The future trajectories of the dynamic obstacles are predicted and divided into segments, and a convex hull is generated for each segment to represent the collision range. Then the flight trajectory, considering yaw direction, is optimized with splines.

In the above methods for obstacle avoidance in dynamic environments, the dynamic obstacles are detected and tracked separately and modeled as certain shapes, which is not favorable when the dynamic obstacle is complex-shaped or is not trained in the detector. The dynamic occupancy map \cite{SMCPHDMap} \cite{RFSMap} \cite{DynamicMapICRA2021} \cite{RNNDynamicMap} is a recently proposed approach that can model arbitrary-shaped dynamic obstacles and static obstacles simultaneously in the map with the consideration of the velocity states. In \cite{OursDynamicMap}, we proposed an efficient particle-based local dynamic occupancy map that can be applied to small-scale robotic systems. This work leverage this map and develops a planner to form a system that can realize efficient obstacle avoidance in unknown dynamic environments.

%the obstacle avoidance methods in dynamic environments can mainly be divided into two categories. The first category utilizes learning-based methods to map the sensing data, such as RGB image or depth image, directly to control commands \cite{UAVLearningObstacleAvoidance1} \cite{SociallyAwareLearning} \cite{SociallyCompliantLearning} \cite{OursLearning}. Obstacle status prediction and planning are implicitly contained in the networks. Although learning-based methods have embraced substantial improvements in recent years, it is still difficult for them to adjust to complicated dynamic environments and generalize to untrained scenes.

%To further represent the free space in the map with a form easy to be processed in the planner, safety corridors \cite{VoxelCorridor} \cite{SikangLiuCorridor} and distances fields \cite{Ringbuffer} \cite{FastPlanner} are calculated.

%This section first presents an overview of the complete obstacle avoidance system in dynamic environments. Then we describe the RAS local planner in detail. Finally, the algorithm to connect the local trajectory to an arbitrary global trajectory is expressed.

\section{System Overview} \label{Section: system overview}
\begin{figure}
\centerline{\psfig{figure=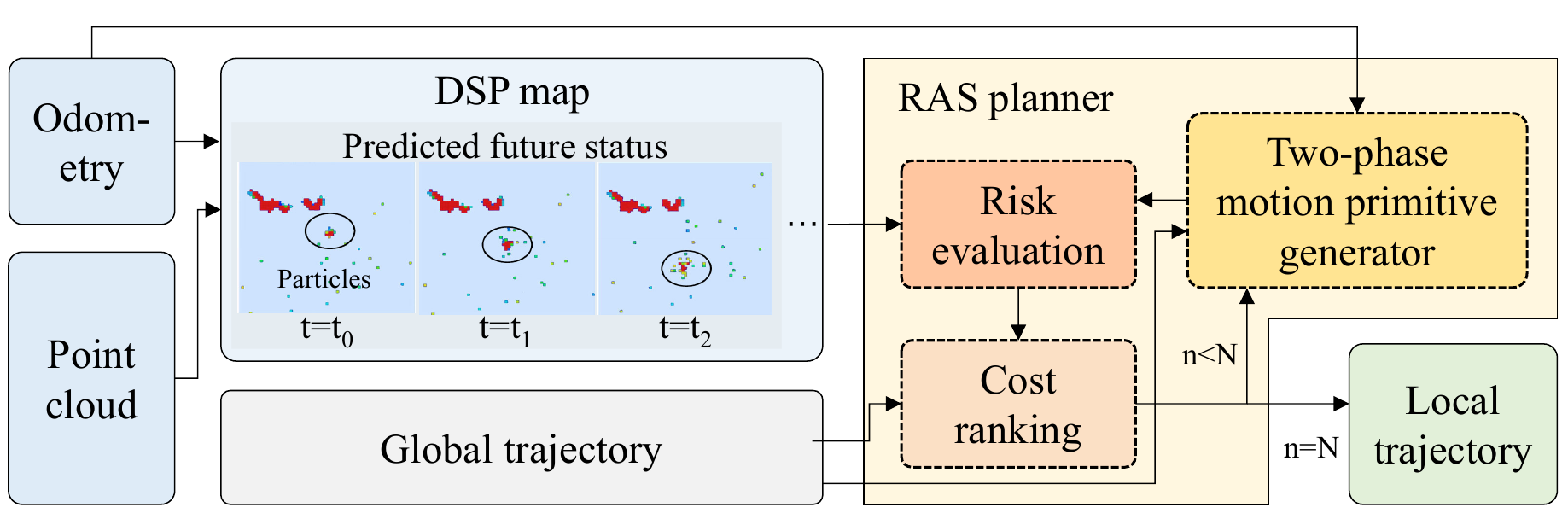, width=3.4in}}
\caption{System structure. The subfigures in the DSP map module show a 2D view of the predicted particle distribution in the future. The particle weight summation is large in a red square and small in a yellow or green square. Particles in the ellipsoid correspond to a dynamic obstacle that is moving to the bottom. The particles disperse as the prediction time increases, which describes the growing variance in predicting the position of the dynamic obstacle. The RAS planner takes the prediction as input and considers risks to generate a local trajectory.}
\label{Figure: Structure}
\end{figure}

The system structure is shown in Fig. \ref{Figure: Structure}. The obstacles, including static obstacles and dynamic obstacles, are sensed in the point cloud form. With the point cloud and the pose of the quadrotor given by a state estimator such as visual odometry, we can build a DSP map \cite{OursDynamicMap}. In this map, particles with velocities are used to estimate the states of obstacles. The occupancy status at a future time can be predicted by propagating the particles and calculating the weight summation. In this work, we utilize the particles in the risk-checking corridor to evaluate the risk of a piece of trajectory with the consideration of the time dimension, which is specified in Section \ref{Section: Risk-aware Sampling}. A safe local trajectory is planned by generating two-phase motion primitives and evaluating their risks. The risks are further regarded as a part of the cost to rank the motion primitives and find the best one.
$N$ pieces of motion primitives, connected smoothly by the junction point of two phases, are used to form the local trajectory.
If no global trajectory is given, the planner adopts a goal position or a goal direction to guide the local trajectory. If a global trajectory is given and is safe temporarily, the planner uses the global trajectory. If an unexpected obstacle occurs and the global trajectory is not safe, the planner adopts the local trajectory and merges back to the global trajectory later. Details of the DSP map can be found in \cite{OursDynamicMap}. The following describes the RAS planner and the algorithm to fuse the local trajectory to an arbitrary global trajectory.

\section{Risk-aware Sampling} \label{Section: Risk-aware Sampling}
In this section, we first introduce how to calculate the risk. Then the trajectory sampling considering the risk is described. Finally, the approach to fuse to the global trajectory is expressed.

\subsection{Risk Evaluation}  \label{Section: Risk Calculation}
In the DSP map \cite{OursDynamicMap}, the obstacles are considered as point objects. Under Gaussian assumption, the point objects are estimated with particles using the sequential Monte Carlo probability hypothesis density (SMC-PHD) filter. Since the environment is unknown and dynamic, the number and the states of the point objects are both random, and thus the point objects form a random finite set (RFS). In a subspace $\mathbb{S}^V$ in the map, the inside point objects can also form an RFS. Let $O$ denote this RFS. The cardinality expectation, which reflects the estimated number of point objects, of $O$ is given by
\begin{equation}\label{Eq: cardinality}
  E\left[ |O| \right]= \sum_{\tilde{x}^{(i)}\in  \mathbb{S}} w_i
\end{equation}
where $\tilde{x}^{(i)}$ is the state, containing 3D position and 3D velocity, of particle $i$ and $w_i $ is the corresponding weight. $E[\cdot]$ is the expectation and $|\cdot|$ is the cardinality.
%A particle is represented with $w_i \delta(x-\tilde{x}^{(i)})$, where $\delta(\cdot)$ is the Dirac function.

In the prediction step of the DSP map \cite{OursDynamicMap}, the states of particles are predicted with a mixture model that considers a constant velocity model with Gaussian noise for dynamic obstacles and a static model for static obstacles.
Suppose at a future time $t$, the state of a particle can be predicted as $\tilde{x}^{(i)}(t)$. Then the cardinality expectation of $O$ at time $t$ can be written as:
\begin{equation}\label{Eq: cardinality with t}
E\left[ |O(t)| \right]= \sum_{\tilde{x}^{(i)}(t)\in  \mathbb{S}} w_i  \equiv  E^\prime(\mathbb{S}, t)
\end{equation}
where $E^\prime(\cdot)$ is a function defined to represent the cardinality expectation in the form related to subspace $\mathbb{S}$ and time $t$. %$O(t)$ is the RFS composed of the point objects in $\mathbb{S}$ at time $t$.

With the Gaussian assumption, the uncertainty of the predicted positions of point objects increases with time $t$. At the particle level, the predicted particles disperse, and the cardinality expectation in a small space, such as a traditional voxel space, can be small even if the space is near the distribution center of a predicted position. Thus determining the future occupancy status of a voxel in binary and conducting traditional collision checking is not suitable. Our risk evaluation approach considers the spatial-temporal integral along with a complete piece of trajectory.

\begin{figure}
\centerline{\psfig{figure=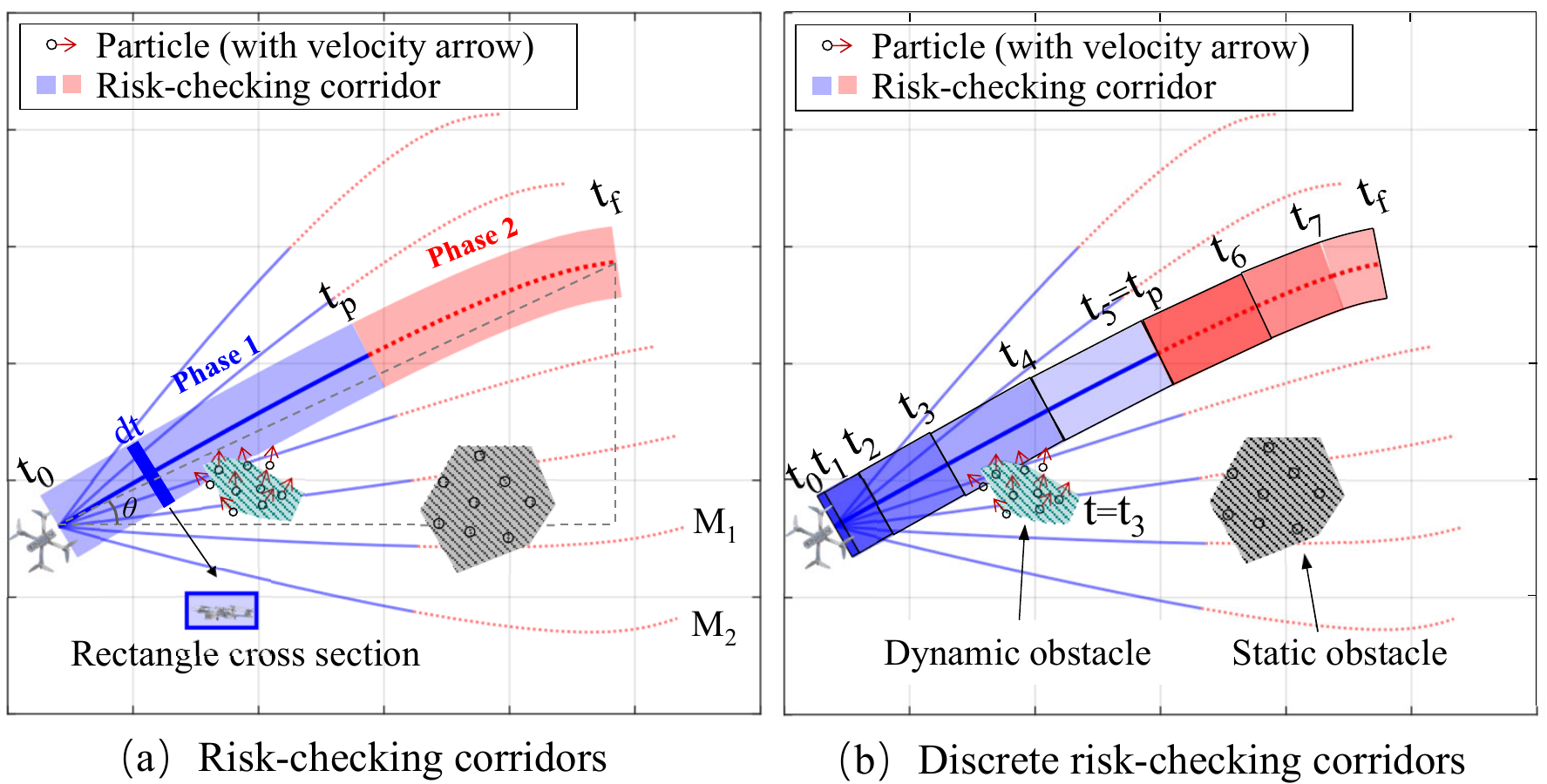, width=3.4in}}
\caption{Risk checking for each motion primitive. Subplot (a) shows the risk-checking corridor used to continuously calculate the risk of two phases of a motion primitive. Subplot (b) shows the risk-checking corridor used for discrete calculation. The grey polygon shows a static obstacle. The green polygon shows a dynamic obstacle. The static obstacle and the dynamic obstacle are represented with particles simultaneously in the DSP map. At $t=t_3$, the dynamic obstacle approaches approaches the risk-checking corridor from $t_3$ and $t_4$ and thus the risk is high.}
\label{Figure: risk checking}
\end{figure}

Let $\boldsymbol f(t)=\left\{\boldsymbol p(t),  \dot{\boldsymbol p}(t), \ddot{ \boldsymbol p}(t), ...\right\}$ denote a piece of trajectory, where $t\in [0, T]$ and $\boldsymbol p(t)$ is the 3D position vector at time $t$.
Suppose the space that the quadrotor flies through along this trajectory is a corridor space, which is called the risk-checking corridor.
The semi-transparent blue and red ribbons in Fig. \ref{Figure: risk checking} (a) show a 2D view of the risk-checking corridors for two phases on a motion primitive. Each phase can be regarded as a piece of trajectory and corresponds to a corridor.
The cross-section of the corridor is a rectangle that acts as the envelope of the quadrotor. Suppose the length and width of the rectangle are $l$ and $w$, respectively.
Let $d \mathbb{S}^{tu}(t)$ denote the corridor space that the quadrotor flies through during a short time interval from $t$ to $t+dt$.
$d \mathbb{S}^{tu}(t)$ is a cuboid space with size $ \left\{ | \dot{\boldsymbol p}(t)| dt, \ l, \ w \right\}$ (with $|\cdot|$ indicating vector length) and center point $\boldsymbol p(t)$.
%The orientation of the cuboid is determined by $\dot{\boldsymbol p}(t)$. Therefore, $d \mathbb{S}^{tu}$ is a function of $t$ given a planned trajectory $f(t)$ and should be written as $d \mathbb{S}^{tu}(t)$.
%the volume $ds^{tu}$ of the corridor space $d \mathbb{S}^{tu}$ from time $t$ to $t+dt$ is:
%\begin{equation}\label{Eq: corridor volume}
%  ds^{tu} = A | \dot{\boldsymbol p}(t)| dt
%\end{equation}
%where $|\cdot|$ denotes the vector length and $d$ is the differential symbol.

We define the risk at point $\boldsymbol p(t)$ from time $t$ to $t+dt$ as $E^\prime(d \mathbb{S}^{tu}(t), t)$. Then the risk of the trajectory $\boldsymbol f(t)$ is calculated by
\begin{equation}\label{Eq: risk continuous}
  R\left[\boldsymbol f(t) \right] = \int_0^T E^\prime \left[ d \mathbb{S}^{tu}(t), t \right] dt
\end{equation}
which is the spatial-temporal integral of the weights of particles in the corridor along the $\boldsymbol f(t)$ with value $R\left[ \boldsymbol f(t) \right] \geq 0$. The physical interpretation of $R\left[ \boldsymbol f(t) \right]$ is the number of expected point objects that the quadrotor will encounter when it flies along this trajectory. Since the point objects represent the obstacles \cite{OursDynamicMap}, the ideal situation is that $R\left[ \boldsymbol f(t) \right] = 0$. However, in consideration of the uncertainty in the prediction and the inevitable noise in the map, regarding $R\left[ \boldsymbol f(t) \right] = 0$ as a safe trajectory is too conservative and $R\left[ \boldsymbol f(t) \right] < \delta$, where $\delta$ is a small positive value, is taken.

To improve the computational efficiency, the risk calculation is conducted in a discrete form, which is illustrated in Fig. \ref{Figure: risk checking} (b). The trajectory is divided into segments by $\Delta t$. Considering (\ref{Eq: cardinality with t}), Equation (\ref{Eq: risk continuous}) turns to
\begin{equation}\label{Eq: risk discrete}
  R\left[\boldsymbol f(t) \right] = \sum_{j=0}^{N-1} E^\prime \left[ d \mathbb{S}^{tu}(t_j), t_j \right] =  \sum_{j=0}^{N-1} \sum_{\tilde{x}^{(i)}(t)\in  d \mathbb{S}^{tu}(t_j)} w_i
\end{equation}
where $N$ is the number of segments and $t_j = t_0 + j  \Delta t$. $\mathbb{S}^{tu}(t_j)$ is the risk-checking corridor from $t_j$ to $t_{j+1}$.

\subsection{Trajectory sampling} \label{Section: Trajectory sampling}
The local trajectory is composed of several pieces. Each piece is sampled with motion primitives in the state space. Fig. \ref{Figure: risk checking} shows the motion primitives during one sampling procedure. Only a few motion primitives in 2D space are shown for clearness. The motion primitives are generated uniformly to different direction angles \cite{ActiveSensing} \cite{StateSpaceSampling}. Each motion primitive is calculated with the same method presented in \cite{OursPlanning}, which is a computationally efficient approach and considers both jerk optimality and time optimality. Assume the flight time for a primitive is $t_f - t_0$. We divide the motion primitive into two phases with a time parameter $t_p$. Phase 1 describes a short term from $t_0$ to $t_p$ and Phase 2 describes a relatively long term from $t_p$ to $t_f$.

Let $\boldsymbol  f^M_1(t)$ and $\boldsymbol  f^M_2(t)$ denote the trajectory in Phase 1 and Phase 2 of a motion primitive, respectively. Phase 1 is used as a piece of the final local trajectory. It is regarded collision-free only when $R\left[ \boldsymbol f^M_1(t) \right] < \delta$. For the motion primitives with a collision-free Phase 1, we rank them with the following cost
\begin{equation}\label{Eq: cost total}
  J = \lambda_1 J_{r} +  \lambda_2 J_{g} + \lambda_3 J_{d}
\end{equation}
where $\lambda_i$ is positive coefficients and $\sum \lambda_i = 1$.
$J_{r}$ is the risk cost and is given by
\begin{equation}\label{Eq: cost risk}
   J_{r} = R\left[ \boldsymbol f^M_1(t) \right] + R\left[ \boldsymbol f^M_2(t) \right]
\end{equation}
With $J_{r}$, the motion primitive that leads to a low-risk region in a relatively long term is preferred. This enhances the safety in dynamic environments and accelerates the process to find a feasible local trajectory. For example, in Fig. \ref{Figure: risk checking} (a), two motion primitives $M_1$ and $M_2$ are both collision-free in Phase 1. However, $M_1$ leads to a position close to an obstacle. The next piece of motion primitive will either collide or make a large turn. Thus $M_2$ is preferred so that the next piece can be found quickly. Extending the sampling distance and making the whole motion primitive longer can also solve the problem but makes it hard to find a safe primitive in a dense environment.

$J_{g}$ is the cost to get close to the goal position. Suppose the goal position is $\boldsymbol p_g$ and the position at $t$ on the motion primitive is $\boldsymbol p(t)$.
\begin{equation}\label{Eq: goal}
   J_{g} = |\boldsymbol p(t_p)  - \boldsymbol p_g |
\end{equation}
The last cost $J_{d}$ is the cost to reduce direction turning between two pieces of motion primitives. Larger direction turning leads to more energy costs.  Suppose the direction angle for the last chosen motion primitive is $\theta_l$, and the direction angle to the valued motion primitive is $\theta$. $J_d$ is defined as
\begin{equation}\label{Eq: goal}
   J_{d} = (\theta - \theta_l)^2
\end{equation}
In the 3D space, the direction change of vertical angle is also considered.

\begin{figure}
\centering
\framebox{\parbox{3.3in}{
\setlength{\parindent}{0.1in}
\textbf{Input:}
Current state $\boldsymbol s_c$.
 \textbf{Output:}
Trajectory $\boldsymbol f^L(t)$.
%
%\textbf{Parameters:}
%$\delta \overline{\theta}$, $k_{max}$, $\delta \theta_{max}$, $\lambda_1$, $\lambda_2$, $\lambda_3$
%
%\textbf{Variables:}
%$\theta_{small}$, $\theta_{large}$, $\theta^{'}_{small}$, $\theta^{'}_{large}$, $\delta \theta$, $k$

\setlength{\parindent}{0.0in}

\begin{algorithmic}[1]
\State Let $k \gets 1$ and $\boldsymbol s_k \gets  \boldsymbol s_c$, where $\boldsymbol s_c$ is the current state of the quadrotor.
\State Generate two-phase motion primitives $M_k^i$, where $i=1,2,...,I_k$. $C_k$.clear().
\label{step: motion primitive}
\State Calculate risk $R[\boldsymbol f^{M_k^i}_1(t)]$ and $R[\boldsymbol f^{M_k^i}_2(t)]$ for $M_k^i$. \textbf{If} $R[\boldsymbol f^{M_k^i}_1(t)]< \delta$, $C_k$.add($M_k^i$).
\label{step: calculate risk}
\State \textbf{If} $C_k$ is empty, \textbf{goto} \ref{step: check step num}. \textbf{Else}, rank candidates in $C_k$ with the cost in (\ref{Eq: cost total}), select the best motion primitive $M_k^*$ in $C_k$ and \textbf{goto} \ref{step: last check}.
\label{step: check and rank}
\State \textbf{If} $k=1$, \textbf{goto} \ref{step: motion primitive}. \textbf{Else}, $C_{k-1}$.erase($M_{k-1}^*$). Let $k\gets k-1$, and \textbf{goto} \ref{step: check and rank}.
\label{step: check step num}
\State \textbf{If} $k<K$,  $\boldsymbol s_{k+1} \gets \hat{\boldsymbol s}_{k}$, where $\hat{\boldsymbol s}_{k}$ is the state of the junction point between two phases of $M_k^*$, $k \gets k+1$ and \textbf{goto} \ref{step: motion primitive}. \textbf{Else}, connect $f^{M_k^i}_1(t), i=1,2,...,I_k$ to get  $\boldsymbol f^L(t)$.
\label{step: last check}
\end{algorithmic}
}}
\caption{The algorithm to sample the local trajectory.}
\label{Fig: Sample algorithm}
\end{figure}

Phase 1 of the motion primitive with the lowest cost is used as a piece of the local trajectory. The algorithm to find $K$ pieces is presented in Fig. \ref{Fig: Sample algorithm}. A candidate list is used for each piece to rank the valid candidates, which is not safe currently or leads to a state that the next piece cannot be found. In each motion primitive, the trajectory is a five-order polynomial and is thus continuous on the snap level. The junction point between two pieces of motion primitives considers continuity on the acceleration level. Thus the whole trajectory is $G^2$ continuous.
%Since the dynamic environment changes rapidly, we don't consider the optimality globally but value more about the efficiency. In each piece, we consider the objective to reach the goal position quickly and safely by using the costs.

In the dynamic environment, new dynamic obstacles can appear in the map suddenly, and the prediction result of the future map occupancy status changes in real-time. We replan the $k>1$ pieces in real-time at a high frequency to find a trajectory that fits the latest prediction result, which means the algorithm in Fig. \ref{Fig: Sample algorithm} is conducted from $k=2$ in real-time after the Piece $k=1$ is generated. Piece $k=1$ is not replanned unless $R[\boldsymbol f^{M_k^i}_1(t)]< \delta$ is not satisfied, in which case a new dynamic obstacle intrudes closely to $\boldsymbol f^{M_k^i}_1(t)$. This intrusion is usually caused by the limited field of view of the sensors.
%Some dynamic obstacles may be included in the field of view only when they are very close to the quadrotor. Piece $k=1$ is not replanned in real-time because, with the uncertainty of the future status of dynamic obstacles, the motion primitive with the smallest cost sometimes swings back and forth and leads to a position too close to the dynamic obstacles that a collision cannot be avoided.
When the quadrotor finishes flying along Piece $k=1$ without replanning, Piece $k=n$ replaces Piece $k=n-1$ so that the trajectory is continuous. To further reduce the response time for the intrusion, $I_k$ is small, indicating a coarse sampling, is taken when $k=1$ in \ref{step: motion primitive} in Fig. \ref{Fig: Sample algorithm}. When $k>1$, $I_k$ can be large to get a refined sampling result.
In practice, $K=2$ is taken because the sensing range of our quadrotor is limited, and the long-term prediction of the dynamic environments is not accurate.

\subsection{Global Trajectory Fusion}  \label{Section: Global Trajectory Fusion}
The above planner considers planning a local trajectory. In some tasks, such as the patrol task and the transportation task, a global trajectory is usually planned in a previously constructed static map. The quadrotor is supposed to follow the global trajectory unless a new obstacle or a dynamic obstacle blocks the trajectory. Therefore, we present the approach to fusing our local trajectory with the global trajectory in this section. The global trajectory is supposed to be arbitrary and is composed of many trajectory points, including 3D position, velocity, and acceleration. The idea of fusion is quite intuitive. When the global trajectory is not safe, the quadrotor flies with the local trajectory. Then the quadrotor merges back to the global trajectory when the global trajectory turns safe.

\begin{figure}
\centering
\framebox{\parbox{3.3in}{
\setlength{\parindent}{0.1in}
\textbf{Input:}
A global trajectory with position points $\left\{ \boldsymbol p_g^1, \boldsymbol p_g^2, ... , \boldsymbol p_g^j, ... , \boldsymbol p_g^J \right\}$.

\setlength{\parindent}{0.1in}
 \textbf{Output:}
Key points vector $P_{k}$.
%
%\textbf{Parameters:}
%$\delta \overline{\theta}$, $k_{max}$, $\delta \theta_{max}$, $\lambda_1$, $\lambda_2$, $\lambda_3$
%
%\textbf{Variables:}
%$\theta_{small}$, $\theta_{large}$, $\theta^{'}_{small}$, $\theta^{'}_{large}$, $\delta \theta$, $k$

\setlength{\parindent}{0.0in}

\begin{algorithmic}[1]
\State $P_{k}$.add($\boldsymbol p_g^1$).  $P_{k}$.add($\boldsymbol p_g^{n_{s1}}$).
\State $j \gets n_{s1}$, $c \gets 0$, $m \gets 2$.
\While{$j<J$}
\State  $j \gets j+1$, $c \gets c+1$
\State  \textbf{If} $\left \langle P_{k}(m)-P_{k}(m-1), \ \boldsymbol p_g^j - P_{k}(m) \right \rangle > \delta \theta$: find the point $\boldsymbol p_g^h \in \left\{ P_{k}(m), ...,  \boldsymbol p_g^j \right\}$ that is the farthest point to line $P_{k}(m)$-$\boldsymbol p_g^j$.
\State $P_{key}$.add($\boldsymbol p_g^h$). $m \gets m+1$. $c \gets 0$ $j \gets h$.
\State  \textbf{Else if} $c>C$: $P_{key}$.add($\boldsymbol p_g^j$). $m \gets m+1$. $c \gets 0$.
\State  \textbf{Else}: \textbf{continue}.
\EndWhile
\State $P_{key}$.add($\boldsymbol p_g^J$).
\end{algorithmic}
}}
\caption{The algorithm to select key points on the global trajectory.}
\label{Fig: key point algorithm}
\end{figure}

\begin{figure}
\centerline{\psfig{figure=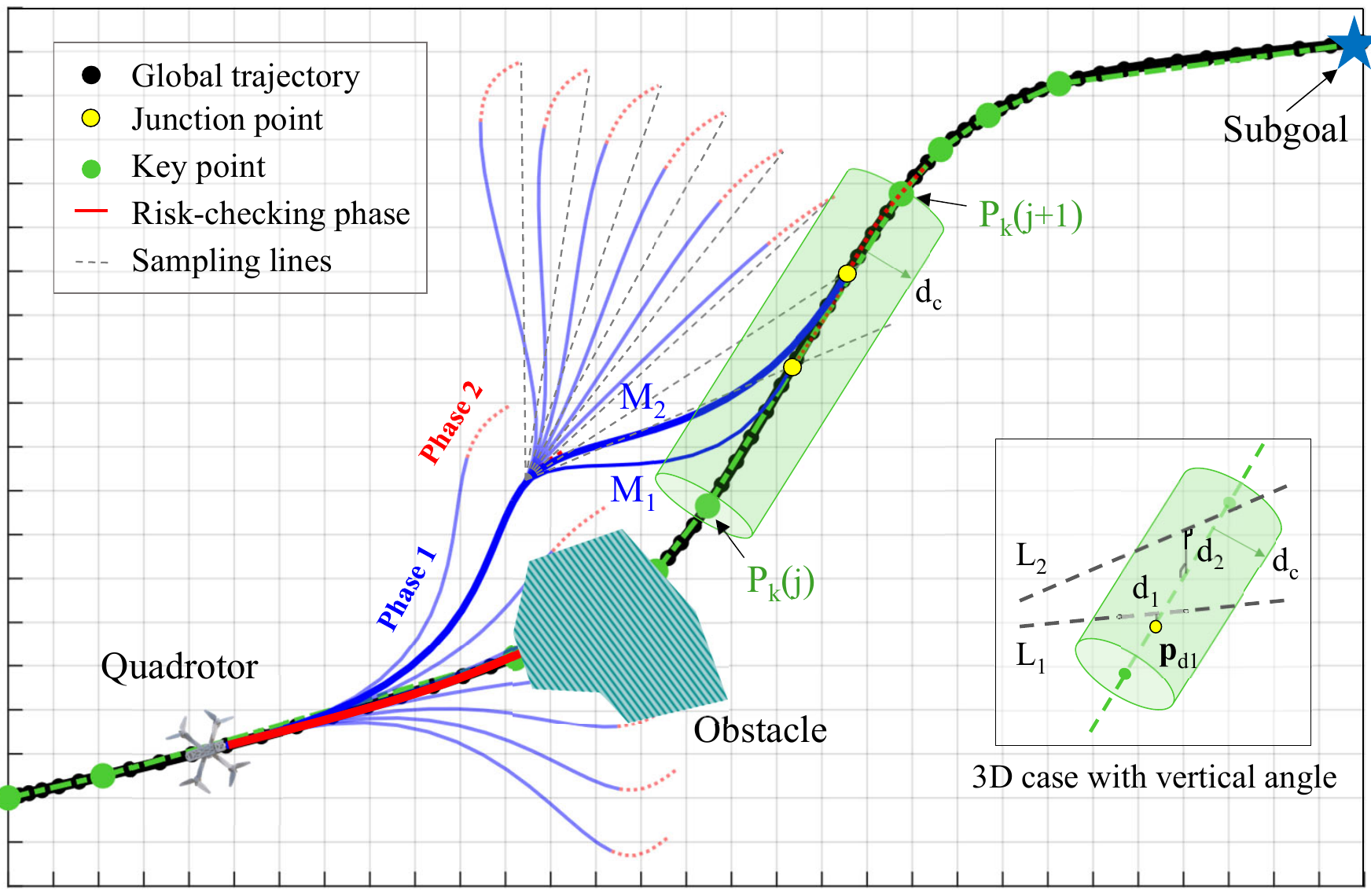, width=3.4in}}
\caption{Illustration of the fusion of local trajectory and global trajectory. The black curve shows a preplanned global trajectory. When the global trajectory is not safe (determined by the risk of the trajectory in the risk-checking phase), a safe local trajectory is planned.
The global trajectory is approximated by polylines composed of the searched key points (green). Sampling line segments (gray dashed lines) are generated before the motion primitive to check if the motion primitive can connect back to the global trajectory. If the condition that the distance between the sampling line segment to one polyline is smaller than $d_c$ is satisfied, the end state of the corresponding motion primitive is chosen as a point on the global trajectory to realize the merging. The green cylinders with radius $d_c$ show the range to determine the condition.
In this figure, motion primitives $M_1$ and $M_2$ can merge to the global trajectory. The small window on the right shows a 3D view of two sampling line segments, $L_1$ and $L_2$. The distance between $L_1$ and the polyline is smaller than $d_c$, while the distance between $L_2$ and the polyline is not.
 }
\label{Figure: global fusion}
\end{figure}

Firstly, we select key points on the global trajectory points to approximate the path of the global trajectory with polylines. These polylines are later used to check if a motion primitive in the local planner can connect to the global trajectory.
The algorithm to select the key points is shown in Fig. \ref{Fig: key point algorithm}, where $n_{s1}$ is a parameter that selects the second key point.
$m$ is the number of key points that are found already, $j$ is the current searching point number, and $c$ is the number of points between the current searching point and the latest key point. $C$ is a parameter that controls the maximum number of trajectory points between two key points. The green dots in Fig. \ref{Figure: global fusion} show an example of the key points. With our algorithm, the key points are dense near the trajectory with a large curvature and sparse near the trajectory with a small curvature.

When the quadrotor is flying with the global trajectory, a risk-checking phase composed of a series of trajectory points ahead is used to determine the risk of the global trajectory. The risk-checking phase is shown with the red curve in Fig. \ref{Figure: global fusion}. Using the same condition as Phase 1 in Section \ref{Section: Risk-aware Sampling}, if the risk is smaller than $\delta$, a new local trajectory is planned with the local planner in Section \ref{Section: Risk-aware Sampling} and the quadrotor flies with the local trajectory. To merge the local trajectory back to the global trajectory safely and smoothly, the junction points between the local trajectory and the global trajectory need to be selected, and the cost for the motion primitives needs to be modified.

The motion primitives in Section \ref{Section: Risk-aware Sampling} are sampled uniformly to different directions with a certain distance. We generate a sampling line segment $L$ to each direction before calculating the motion primitive. The gray dashed lines in Fig. \ref{Figure: global fusion} show the sampling line segments. If the distance $d$ between $L$ and the line segment defined by $P_k(j)$ and $P_k(j+1)$ is shorter than a threshold $d_c$, the endpoint of the motion primitive is moved to the point, $\boldsymbol p_d$, closest to $L$ on the line segment $P_k(j)-P_k(j+1)$. Then the motion primitive merges to the global trajectory at $\boldsymbol p_d$. The calculation of $d$ and $\boldsymbol p_d$ can be referred to the cylinder-cylinder model in \cite{Cylinder-Cylinder}.
It is possible that a feasible motion primitive \cite{OursPlanning} cannot be generated with the position, velocity, and acceleration on the new endpoint because of the limited dynamic performance of the quadrotor. In this case, the endpoint is searched near $\boldsymbol p_d$. If the motion primitive still cannot be generated, the connection for this motion primitive is determined to be unavailable.

In the ranking procedure in the local planner, the cost function becomes:
\begin{equation}\label{Eq: cost total with global}
  J = \lambda_1 J_{r} +  \lambda_2 J_{g} + \lambda_3 J_{d} + \lambda_4 J_{c}
\end{equation}
The goal position to calculate $J_{g}$ turns to a subgoal, which is the intersection of the global trajectory and the boundary of the local map. $J_{c}$ is the cost to select the motion primitives close to the global trajectory and is given by
\begin{equation}\label{Eq: cost global}
  J_{c} = \left\{
  \begin{aligned}
  & 0, & \text{if} \ \ d \leq d_c  \\
  & (d - d_c)^2, & \text{otherwise}
  \end{aligned}
  \right.
\end{equation}

The key points selection procedure can be conducted offline before the flight. The rest of the computations are analytical calculations or querying operations, and thus the planner is computationally efficient. In our tests, the planner takes 1.84 ms at most to plan a local trajectory with an AMD R7-4800H CPU.

\section{Experiments}
The experiments were conducted in both simulation and the real world. In the simulation tests, our planning approach is compared with several other approaches. In the real-world tests, we first utilized the true position data from a motion capture system to maximize the performance of the planner and then adopted onboard sensing and computing of a mini quadrotor to realize autonomous obstacle avoidance in dynamic environments.

\subsection{Simulation Tests}
\begin{figure}
\centerline{\psfig{figure=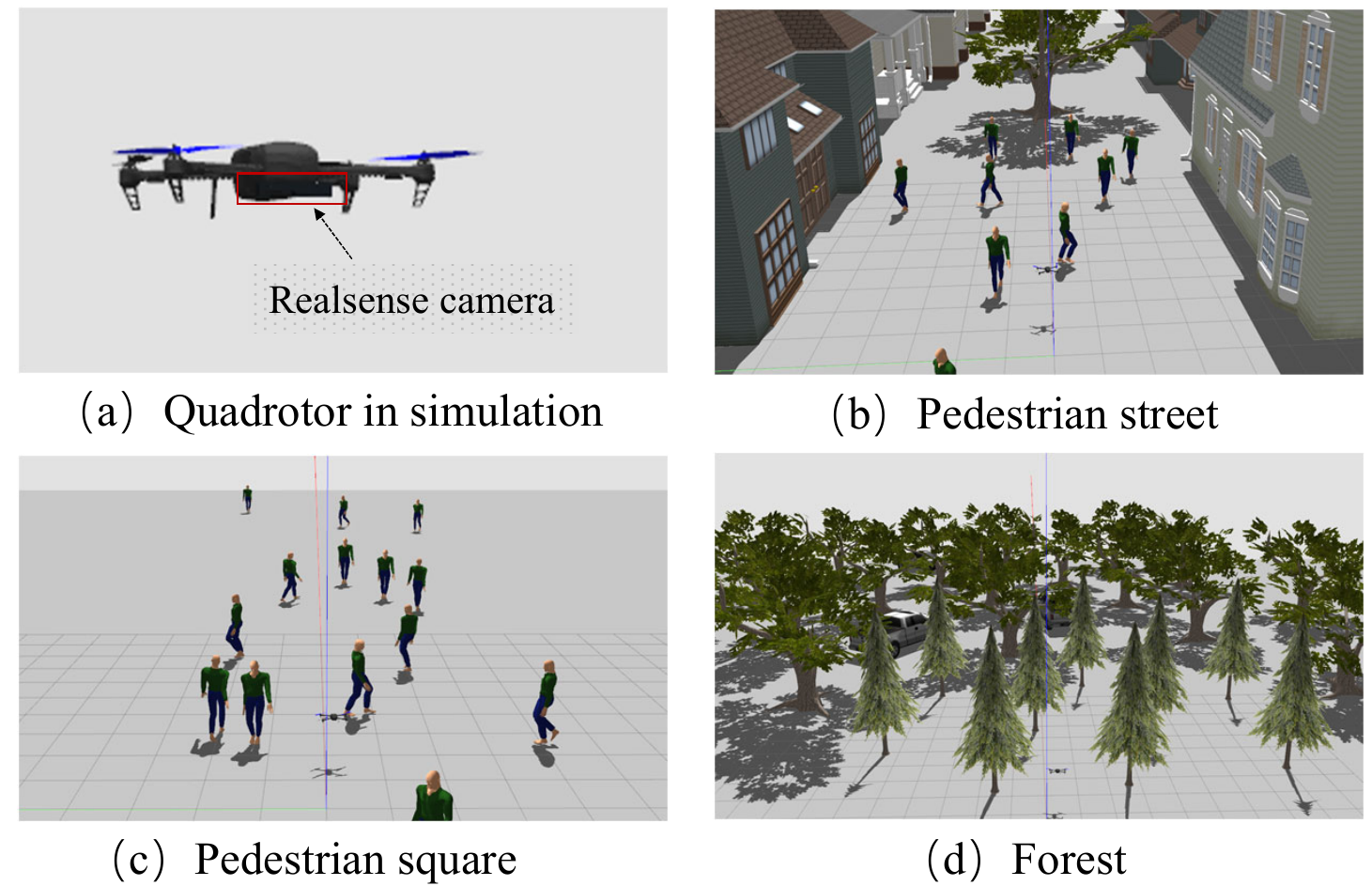, width=3.4in}}
\caption{Quadrotor in the simulation and three testing worlds. (b) is a static environment. (c) and (d) are dynamic environments. }
\label{Figure: simulation_settings}
\end{figure}

The Gazebo simulation environment with the PX4 firmware is adopted in the simulation tests. The quadrotor is equipped with a Realsense camera to get the point cloud of obstacles in the field of view, and the DSP map is built with the point cloud. Then the RAS planner is used to plan the trajectory composed of position, velocity, and accelerate commands. These commands are sent to a PID tracker to control the quadrotor's attitude and thrust. Experiments were conducted in two dynamic worlds and a static world shown in Fig. \ref{Figure: simulation_settings}. We compared our system with three different methods.
The first method \cite{ActiveSensing} adopts the static local map \cite{Ringbuffer} to represent static obstacles and the DATMO pipeline to represent and predict dynamic obstacles separately. Planning is conducted with real-time uniform state-space sampling.
The second method is a computationally efficient planner \cite{OursPlanning}, named Nags planner, for high-speed flights in static environments. The obstacles are represented with a static local map \cite{Ringbuffer} and no prediction for dynamic obstacles are considered.
The third method also uses the DSP map to represent the environment, but only Phase 1 is considered in the planning procedure.

In all the methods, the maximum velocity for planning was $3m/s$, and the maximum acceleration was $4 m/s^2$. Each method was tested 20 times in each map. The results can be found in Table \ref{Table: simulation comparison}. The obstacle avoidance performance is evaluated with three metrics: the total collision times, the total freezing times, and the average flight time. In one test, the quadrotor might collide multiple times. We manually relocate the drone to a nearby safe position when a collision happens and continue the test. When the quadrotor cannot find a safe trajectory, it encounters the freezing robot problem.
Since the quadrotor has a limited field of view, some dynamic obstacles might not be sensed, and absolutely safe flights are difficult to achieve in dynamic environments. Thus in the pedestrian street world and the pedestrian square world, all the methods collide at least one time in the 20 tests. By using the risk-aware sampling, our RAS planner has the least collision times and the shortest flight time in dynamic environments, and the number of freezing is zero. The Nags planner has the shortest flight time in the static forest environment, but the collision to dynamic obstacles frequently happens since it has no prediction. DATMO and sampling method \cite{ActiveSensing} takes the longest flight time while the number of collisions is in the middle. In the static environment, all the planners have collision-free flights except the one-phase planning method. This method considers only the short-term risk and sometimes leads to a position with complex obstacles, such as a small vacancy between branches, and causes a collision. Overall, our RAS planner has the best performance.

We also tested the ability to fuse to different global trajectories in the simulation. The global trajectories are generated using two different approaches. The first approach uses the kino-dynamic A* in \cite{FastPlanner} to search the global trajectory. The trajectory is $G^1$ continuous. The second approach is to adopt multiple motion primitives described in Section \ref{Section: Trajectory sampling} to reach the goal and the generated global trajectory is $G^2$ continuous. Fifty global trajectories are generated with each approach. Our planner successfully replanned a safe local trajectory when the global trajectory was unsafe and merged to the global trajectory later in all the tests.

\begin{center}
\begin{table*}[]
\centering
\caption{Comparison Results in Different Simulation Worlds}

\begin{tabular}{l|ccc|ccc|ccc}
\hline
World                     & \multicolumn{3}{c|}{Pedestrian Street}                                                                                                                                                                                          & \multicolumn{3}{c|}{Pedestrian Square}                                                                                                                                                                                         & \multicolumn{3}{c}{Froest}                                                                                                                                                                                                      \\ \hline
Metric                    & \multicolumn{1}{c|}{\begin{tabular}[c]{@{}c@{}}Collison \\ Times\end{tabular}} & \multicolumn{1}{c|}{\begin{tabular}[c]{@{}c@{}}Freezing \\ Times\end{tabular}} & \begin{tabular}[c]{@{}c@{}}Avg. Flight\\ time (s)\end{tabular} & \multicolumn{1}{c|}{\begin{tabular}[c]{@{}c@{}}Collison \\ Times\end{tabular}} & \multicolumn{1}{c|}{\begin{tabular}[c]{@{}c@{}}Freezing\\ Times\end{tabular}} & \begin{tabular}[c]{@{}c@{}}Avg. Flight\\ time (s)\end{tabular} & \multicolumn{1}{c|}{\begin{tabular}[c]{@{}c@{}}Collison \\ Times\end{tabular}} & \multicolumn{1}{c|}{\begin{tabular}[c]{@{}c@{}}Freezing \\ Times\end{tabular}} & \begin{tabular}[c]{@{}c@{}}Avg. Flight\\ time (s)\end{tabular} \\ \hline
DATMO \& Sampling {[}1{]} & \multicolumn{1}{c|}{4}                                                         & \multicolumn{1}{c|}{2}                                                        & 28.9                                                           & \multicolumn{1}{c|}{6}                                                         & \multicolumn{1}{c|}{0}                                                       & 43.6                                                           & \multicolumn{1}{c|}{0}                                                         & \multicolumn{1}{c|}{0}                                                        & 32.0                                                           \\ \hline
Nags Planner {[}8{]}      & \multicolumn{1}{c|}{10}                                                        & \multicolumn{1}{c|}{0}                                                        & 25.8                                                           & \multicolumn{1}{c|}{11}                                                        & \multicolumn{1}{c|}{0}                                                       & 36.0                                                           & \multicolumn{1}{c|}{0}                                                         & \multicolumn{1}{c|}{0}                                                        & \textbf{21.7}                                                  \\ \hline
Ours One-phase            & \multicolumn{1}{c|}{5}                                                         & \multicolumn{1}{c|}{1}                                                        & 27.9                                                           & \multicolumn{1}{c|}{2}                                                         & \multicolumn{1}{c|}{0}                                                       & 39.8                                                           & \multicolumn{1}{c|}{2}                                                         & \multicolumn{1}{c|}{0}                                                        & 32.5                                                           \\ \hline
Ours RAS Planner          & \multicolumn{1}{c|}{\textbf{1}}                                                & \multicolumn{1}{c|}{0}                                                        & \textbf{22.8}                                                  & \multicolumn{1}{c|}{\textbf{1}}                                                & \multicolumn{1}{c|}{0}                                                       & \textbf{34.8}                                                  & \multicolumn{1}{c|}{0}                                                         & \multicolumn{1}{c|}{0}                                                        & 32.4                                                           \\ \hline
\end{tabular}

\label{Table: simulation comparison}
\end{table*}
\end{center}

\subsection{Real-world Tests}
\begin{figure}
\centerline{\psfig{figure=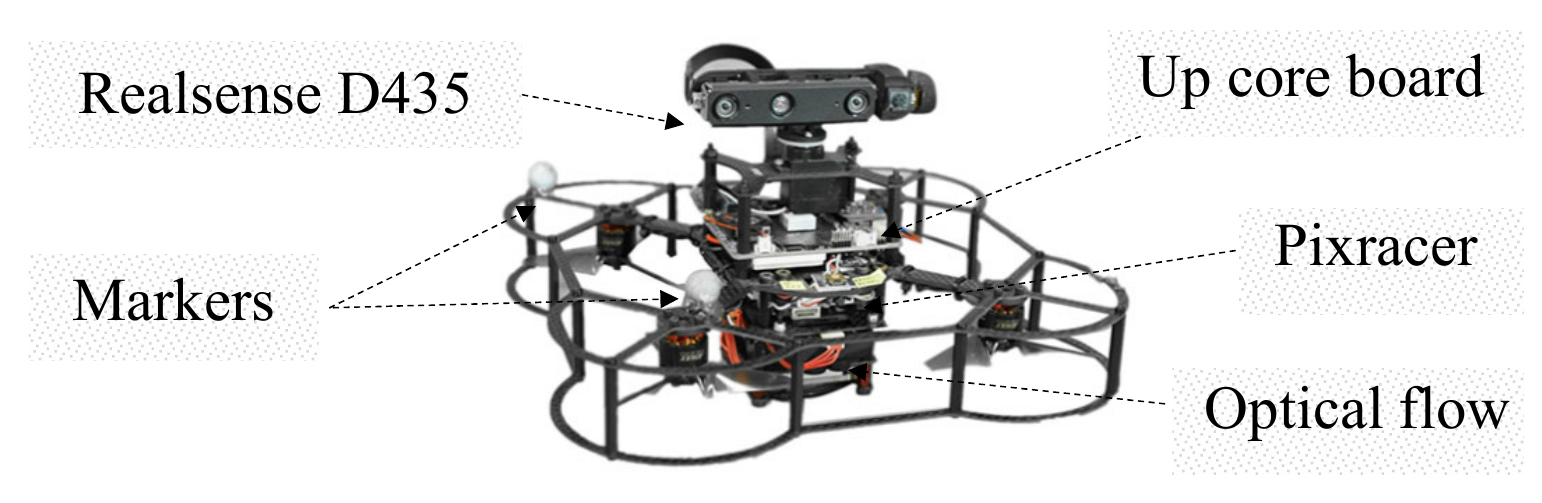, width=2.8in}}
\caption{The hardware structure of the quadrotor.}
\label{Figure: quadrotor}
\end{figure}

In the real-world tests, a quadrotor named Mantis \footnotemark[1] that weighs $320$ grams was used. Fig. \ref{Figure: quadrotor} shows the hardware configuration of the quadrotor. The tests can be divided into two groups. The first group used the true position data from a motion capture system to validate the effectiveness of the planner, while the second group adopted onboard sensing and computing to validate the effectiveness of the entire system.

\subsubsection{Tests with the motion capture system}
\begin{figure}
\centerline{\psfig{figure=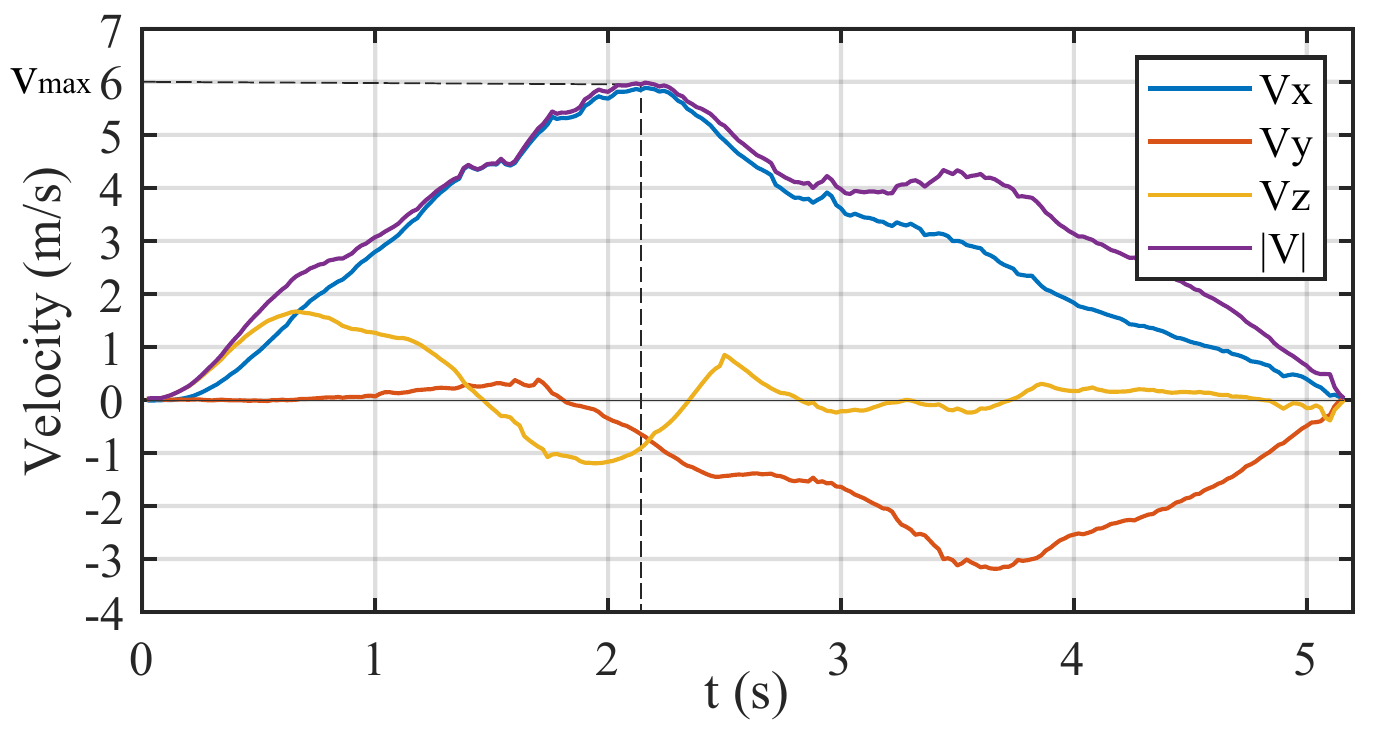, width=3.4in}}
\caption{The velocity curve of a test using the motion capture system.}
\label{Figure:  velocity curve}
\end{figure}

As is shown in Fig. \ref{Figure: realworld_settings} (a), two static obstacles and two dynamic obstacles were in the testing field. The dynamic obstacles moved with a constant velocity, about 1m/s for the faster one and 0.5m/s for the slower one. The Nokov motion capture system was used to capture the position of the quadrotor and the obstacles. The size of the obstacles was pre-measured.
The future occupancy status of the dynamic obstacles was predicted with the constant velocity model under the Gaussian assumption. The quadrotor was able to fly rapidly and safely in the tests with a maximum speed of 6m/s. A corresponding velocity curve is shown in Fig. \ref{Figure:  velocity curve}.

\subsubsection{Tests with onboard sensing and computing}
\footnotetext[1]{See \url{https://www.byintelligence.com}}
\footnotetext[2]{From \url{https://www.cpubenchmark.net/cpu_list.php.}}

\begin{figure}
\centerline{\psfig{figure=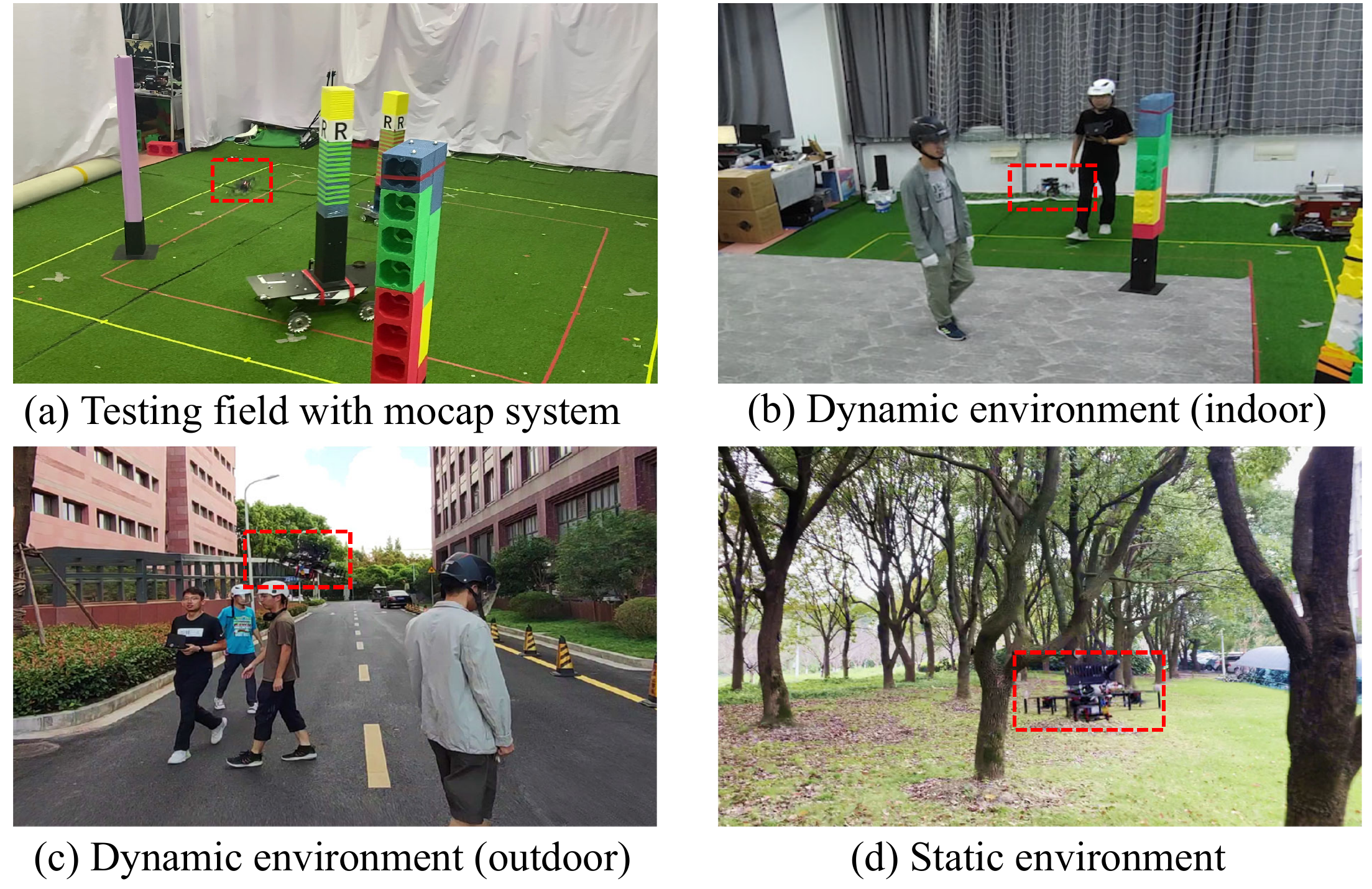, width=3.3in}}
\caption{Testing scenarios in the real world. In subplot (a), the position of the quadrotor and the position of obstacles are given by the motion capture system. The dynamic obstacles are two foam pillars mounted on mobile robots. Subplot (b) and (c) show two dynamic environments. Subplot (d) shows a static environment. }
\label{Figure: realworld_settings}
\end{figure}

\begin{figure}
\centerline{\psfig{figure=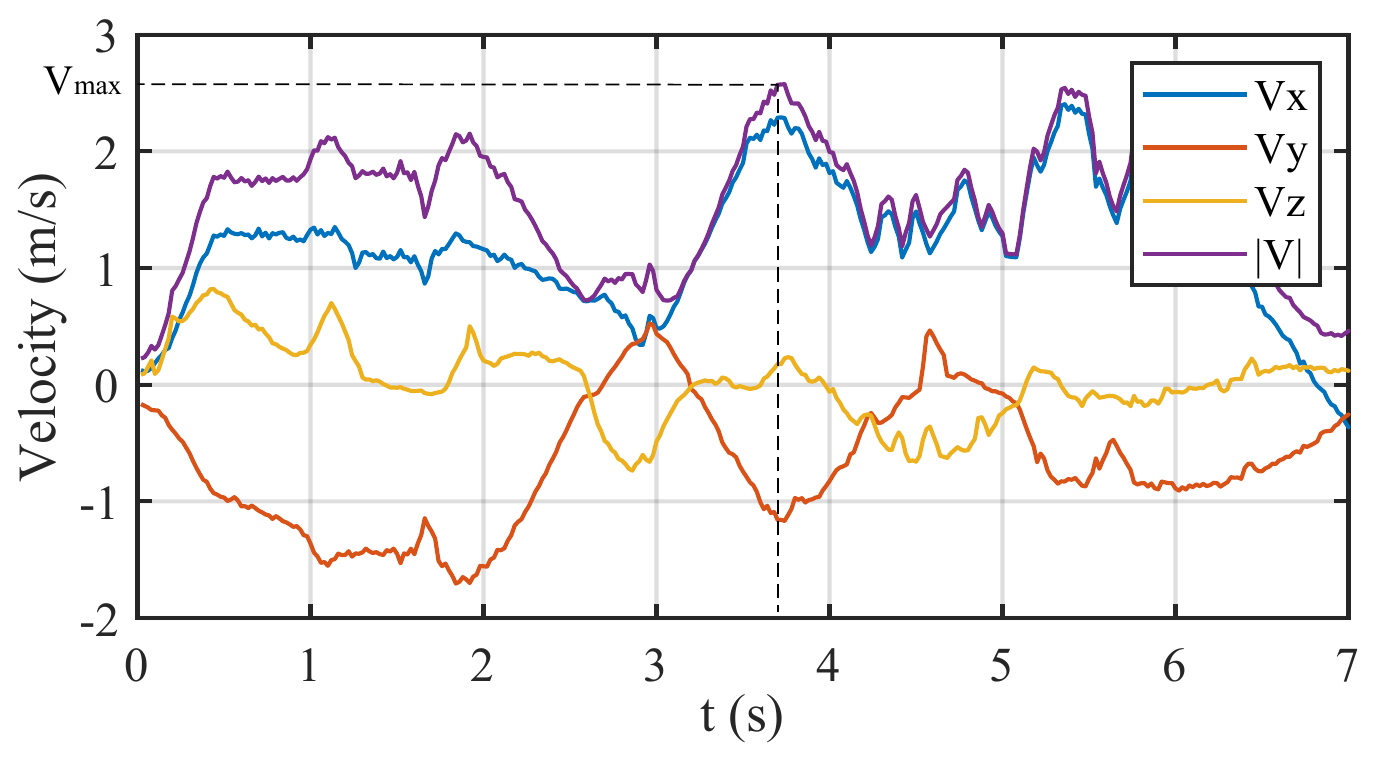, width=3.4in}}
\caption{The velocity curve of a test using onboard sensing and computing in the indoor scenario shown in \ref{Figure: realworld_settings} (b). Velocity estimation from the motion capture system is used and only used to plot the curve.}
\label{Figure:  velocity curve_optical_flow}
\end{figure}

In these tests, the quadrotor utilized an optical flow sensor for positioning, an Intel Realsense D435 depth camera for sensing, and an Up core board to run the DSP map and the RAS planner. The Up core board is a low-price single-board computer that utilizes an Intel atom z8350 processor. The computing power of this processor is only 489 MOps/s \footnotemark[2], which is about one-fifth of the computing power of a regular laptop CPU. Two dynamic environments and one static environment, shown in Fig. \ref{Figure: realworld_settings} (b)-(d), were tested. The dynamic obstacles were pedestrians. The maximum speed is  2.5 m/s in dynamic environments. A velocity curve is shown in Fig. \ref{Figure:  velocity curve_optical_flow}.

\section{Conclusion}
This paper presents a risk-aware sampling-based planner and builds an obstacle avoidance system for dynamic environments. Instead of using DATMO to represent the dynamic obstacles with separate models, we use the DSP map to represent the arbitrary-shaped static and dynamic obstacles simultaneously without detection. Then the risk is defined with cardinality expectation in the predicted maps, and a local trajectory considering the risk can be planned with the RAS planner to avoid collisions. Comparison results in the simulation environments show that our system has the best obstacle avoidance performance in dynamic environments and competitive performance in static environments. In real-world tests, the quadrotor reaches a flight speed of 6 $m/s$ with the motion capture system and 2.5 $m/s$ with everything running on an onboard computer with very low computing power. However, in dynamic environments, the robustness of all the tested systems still has improvement space. One major reason that affects the robustness is that the perception performance with a limited field of view is unsatisfying in an environment with multiple dynamic obstacles.
Future works will investigate perception-aware planning to predict the status of the dynamic obstacles better and enhance robustness of the system.

\bibliographystyle{IEEEtran}
%\enlargethispage{-8.5cm}

\bibliography{asme}

% that's all folks
\end{document}